\documentclass[10pt,final, journal,oneside,letterpaper]{IEEEtran}
\usepackage{CJK}
\usepackage{graphicx}
\usepackage{multirow}
\usepackage{multicol}
\usepackage{stfloats}
\usepackage{setspace}
\usepackage{mathrsfs}
\usepackage{amsmath}
\usepackage{amssymb}
\usepackage{leftidx}
\usepackage{color,soul}
\usepackage{booktabs}
\usepackage[colorlinks,linkcolor=blue]{hyperref}

\begin{document}

\title{FusionLane: Multi-Sensor Fusion for Lane Marking Semantic Segmentation Using Deep Neural Networks}

\author{Ruochen Yin$^{1,2,3}$, Biao Yu$^{1}$, Huapeng Wu$^{3}$, Yuntao Song$^{1}$, Runxin Niu$^{1}$
\thanks{The authors' email addresses are in order as rrolland@mail.ustc.edu.cn, byu@hfcas.ac.cn, Huapeng.Wu@lut.fi, songyt@ipp.ac.cn, rxniu@iim.ac.cn, Runxin Niu and Yuntao Song are the corresponding authors.

1. Hefei Institutes of Physical Science, Chinese Academy of Sciences, Hefei, Anhui, China

2. University of Science and Technology of China, Hefei, Anhui, China

3. Lappeenranta University of Technology (LUT), Lappeenranta, Finland
}
}
\maketitle

\begin{abstract}
It is a crucial step to achieve effective semantic segmentation of lane marking during the construction of the lane level high-precision map. In recent years, many image semantic segmentation methods have been proposed.  These methods mainly focus on the image from camera, due to the limitation of the
sensor itself, the accurate three-dimensional spatial position of the lane marking cannot be obtained, so the demand for the lane level high-precision map construction cannot be met. This paper proposes a lane marking semantic segmentation method based on LIDAR and camera fusion deep neural network. Different from other methods, in order to obtain accurate position information of the segmentation results, the semantic segmentation object of this paper is a bird's eye view converted from a LIDAR points cloud instead of an image captured by a camera. This method first uses the deeplabv3+ [\ref{ref:1}] network to segment the image captured by the camera, and the segmentation result is merged with the point clouds collected by the LIDAR as the input
of the proposed network. In this neural network, we also add a long short-term memory (LSTM) structure to assist the network for semantic segmentation of lane markings by using the the time series information. The experiments on more than 14,000 image datasets which we have manually labeled and expanded have shown the proposed method has better performance on the semantic segmentation of the points cloud bird's eye view. Therefore, the automation of high-precision map construction can be significantly improved. Our code is available at \href{https://github.com/rolandying/FusionLane}{$https://github.com/rolandying/FusionLane$}.
\end{abstract}


\IEEEpeerreviewmaketitle

\section{Introduction}
\IEEEPARstart{A}{s} an important resource in the field of autonomous driving, high-precision map plays a central role. It can not only provide high-precision positioning function based on map matching, but also transmit some complex information of pavement as a priori knowledge to unmanned vehicles, such as lane, slope, curvature, heading, etc. High-precision maps can be seen as an additional complement to the perception module of unmanned vehicles, it helps unmanned vehicles focus on other perceived tasks such as moving obstacles detection and tracking. Therefore, the lane level high-precision map must contain accurate lane marking information.

With the great success of convolutional neural networks (CNN) achieved in the image processing field, a number of CNN-based image semantic segmentation methods have been proposed. Different from the form of the bounding box, pixel-wise prediction result is more in line with the requirements of high-precision map construction. But limited by the camera itself, these methods can not get the accurate spatial position information of the lane marking.
\begin{figure}[hp]
\centering
\begin{minipage}[t]{0.485\linewidth}
\centering
\includegraphics[scale=0.3]{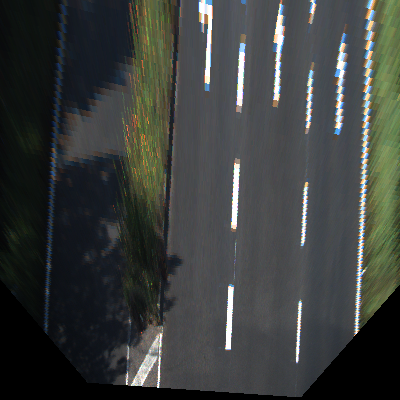}
\end{minipage}%
\begin{minipage}[t]{0.485\linewidth}
\centering
\includegraphics[scale=0.3]{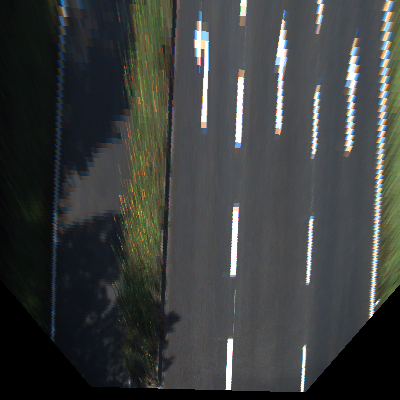}
\end{minipage}
\begin{spacing}{1.1}

\end{spacing}
\begin{minipage}[t]{0.485\linewidth}
\centering
\includegraphics[scale=0.3]{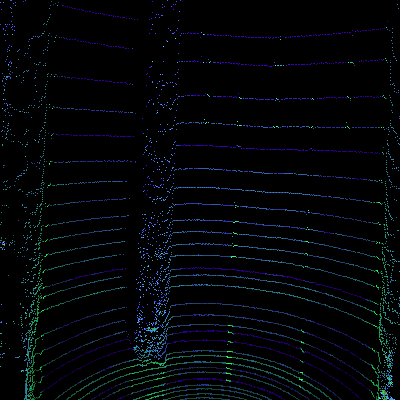}
\end{minipage}%
\begin{minipage}[t]{0.485\linewidth}
\centering
\includegraphics[scale=0.3]{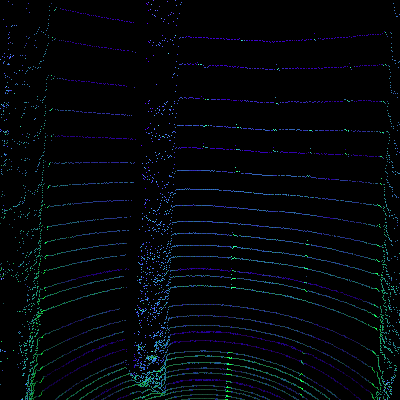}
\end{minipage}%
\caption{These are two consecutive frames of data collected by LIDAR and camera in KITTI datasets. The first row are bird's eye views converted from the front views of the camera. Even if the images are calibrated with the given internal and external parameters of the camera, we can clearly see that the camera bird's eye views are distorted due to the bumpy road surface. Meanwhile, the corresponding LIDAR points cloud bird's eye views in the second row are much more stable. At the same time, the actual physical space corresponding to each pixel becomes larger as the distance extends. On the corresponding bird's eye views, the target details at the top of the image become increasingly blurred. But the LIDAR points cloud bird's eye views do not have this shortcoming.}
\label{fig:1}
\end{figure}
\begin{figure*}[htp]
\centering
\includegraphics[scale=0.22]{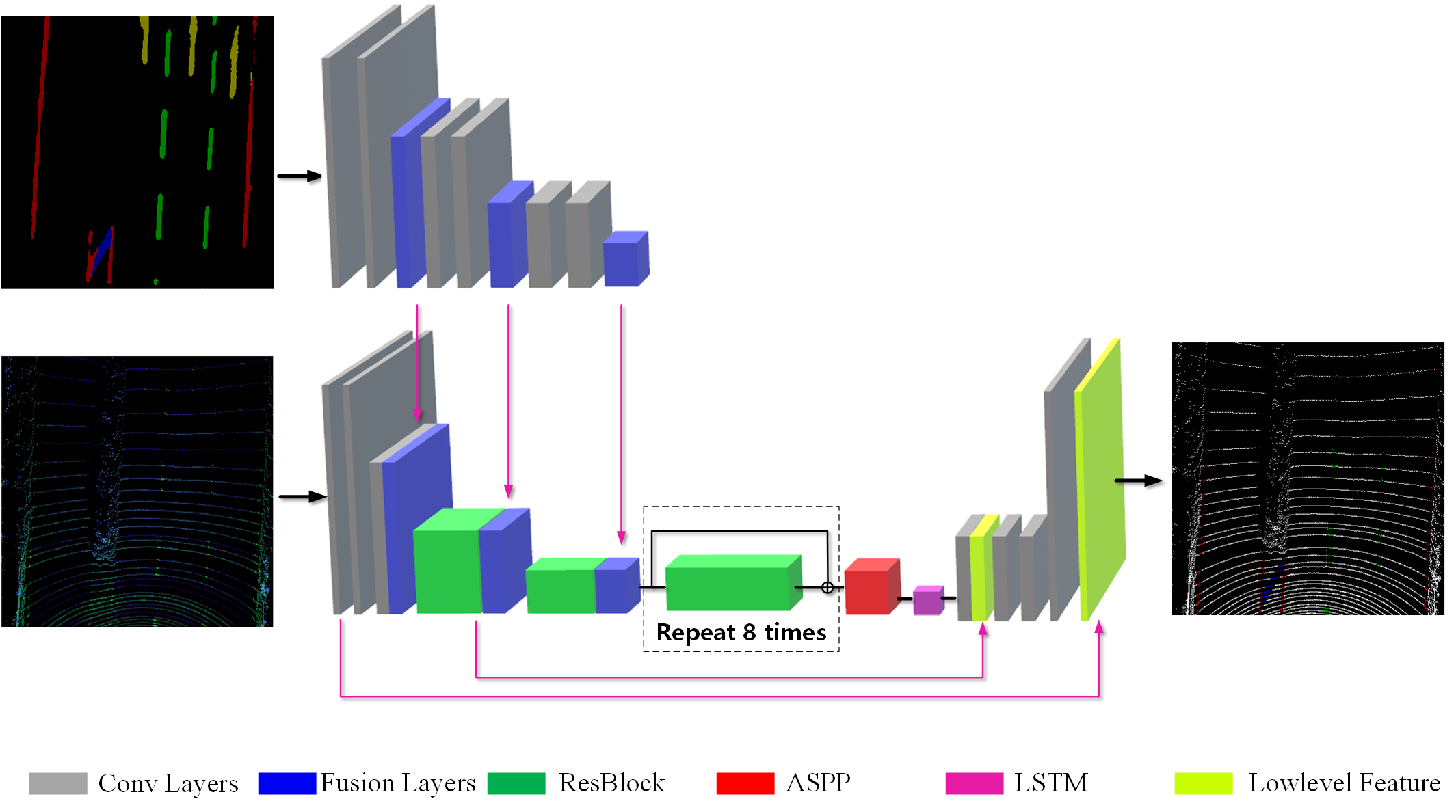}
\caption{The network structure diagram, this diagram mainly shows the general structure of the proposed network, the details will be described below.}\label{fig:2}
\end{figure*}

Some researchers also use the aerial photography for lane marking semantic segmentation [\ref{ref:2}]. The method performs the semantic segmentation of the lane marking by taking the aerial image of the unmanned aerial vehicle (UAV). The advantages are mainly low cost and high efficiency, but the disadvantages are also obvious. First, the classification accuracy of some structural similar elements is not high. Second, the segmentation edge of different types of predictions is not accurate enough. these are mainly because of the aerial photographs contain a lot of irrelevant background information. At the same time, the real space scale corresponding to each pixel in the aerial picture is much larger than the picture collected by the ground platform, which undoubtedly further enlarges the above shortcomings. Therefore, it is difficult to construct a lane-level high-precision map by this method.

In order to solve these problems, the semantic segmentation object for the proposed method is the bird's eye view of the road converted from the LIDAR points cloud (hereinafter referred to as LBEV) instead of a image captured by camera. We propose an encoder-decoder network model which can learn from the visual image and LIDAR points cloud features, and add the LSTM structure to the network to assist the semantic segmentation of the lane marking through timing information. The network basic structure is shown in Fig. 2: First, the front view image acquired by the camera is converted into a bird's eye view, which is semantically segmented by a DeeplabV3+ trained on the corresponding dataset. Then, the obtained result is put into an input branch of the network. After convolution operation, the output feature map of the convolution layer with \emph{convolution stride} = 2 will be fused with the LBEV. Then
the feature map obtained after multiple convolutions is input to the LSTM module as timing information and transmitted to the next moment. Finally, we designed a decoder module which restored the feature map output by the LSTM module to the same size as the original image by two times of bilinear
upsampling. And during the the process of upsampling, we fuse the low level feature from the encoder which makes the decoder better to recover the details of the image.

The main contributions of this paper lie in three-fold.
 \begin{itemize}
 \item[¡¤]First, to the best of our knowledge, this is the first method for the LBEV semantic segmentation, the advantage is that the three-dimensional spatial position of each pixel in LBEV can be accurately obtained, that is to say, the accurate position information of each prediction in the semantic segmentation result can be easily obtained which meet the requirements of high-precision map construction.
 \item[¡¤]Second, based on the KITTI [\ref{ref:3}] datasets, we created a dataset contains more than 14,000 LBEV and camera bird's eye view images(hereinafter referred to as CBEV) each with manual labeling and extension methods.
 \item[¡¤]Third, in practical situations, even after calibration, the pixels in LBEV and CBEV could not be perfectly aligned, so we design an encoder module with two input branches instead of a single input with multiple channels. This allows the network to be independent of the assumption of perfect alignment. In this way, we can fuse the result of CBEV semantic segmentation (hereinafter referred to as C-Region) with LBEV and that makes the segmentation result of the proposed network has both advantages of accurate classification from camera and precise position information from LIDAR. And the LSTM structure can help the network to achieve better prediction results through timing information.
 \end{itemize}

 The remainder of this paper is organized as follows. Section \ref{section:2} mainly reviews the related works. Section \ref{section:3} details the proposed method and network. Section \ref{section:4} compares the proposed method with other methods and analyzes the results. Section \ref{section:5} summarizes the advantages of this method and the future work. And Section \ref{section:6} is the acknowledgement.

\section{Related works}\label{section:2}
In autonomous driving, high-precision map has become an indispensable part. However, building a high-precision map is a difficult and complicated task. Improving the automaticity of the high-precision map construction process and reducing the proportion of human participation have always been the
goal of the researchers. This requires the algorithm not only to infer semantic information from the input image (the need for scene understanding: process from specific to abstraction) but also be able to make pixel-wise segmentation for each category of target (the need for map construction accuracy: process from abstraction to specific), and semantic segmentation meets these requirements.

 Before deep learning was applied to the field of computer vision, researchers generally used Texton Forest or Random Forest method to construct classifiers for semantic segmentation. These methods solved the problem to some extent. But deep learning has revolutionized the field, many computer vision problems, including semantic segmentation, have begun to using the methods based on deep learning frameworks (basically convolutional neural networks), and the effect achieved far exceeds the traditional method. Therefore, only the research status of semantic segmentation based on deep learning framework is introduced below.

\subsection{Semantic segmentation}
At present, all neural networks successfully used for semantic segmentation come from the same work which is the fully-convolutional neural network (FCN) [\ref{ref:4}]. The authors converted some well-known network frameworks such as AlexNet [\ref{ref:5}], VGG-16 [\ref{ref:6}], GoogLeNet [\ref{ref:7}], and ResNet [\ref{ref:8}] into a fully-convolutional structure, replacing the original fully connected layer in these network frames with the some small scale upsampling layers. Semantic segmentation tasks require the network to have the following two capabilities: first, being able to learn multiple scales of feature in the image, second, accurately restoring the details of the original image, especially the edge of the segmentation. To meet these two requirements, researchers
have mainly improved their network in following aspects on the basis of FCN.

\subsubsection{Encoder Variants}
In order to solve the first problem, some models[\ref{ref:9}][\ref{ref:10}] resize the input for several scales and fuse the features from all the scales. [\ref{ref:11}] who transform the input image through a Laplacian pyramid, feed each scale input to a deep convolutional neural network (DCNN) and merge the feature maps from all the scales. [\ref{ref:12}][\ref{ref:13}] employs spatial pyramid pooling to capture context at several ranges. DeeplabV2 [\ref{ref:14}] proposes atrous spatial pyramid pooling (ASPP), where parallel atrous convolution layers with different rates capture multi-scale information.
\subsubsection{Decoder Variants}
Some researchers take different upsampling methods in the decoder module, the purpose is mainly to improve the ability of the decoder to restore the details of the original picture. [\ref{ref:4}][\ref{ref:15}] employ deconvolution [\ref{ref:16}] to learn the upsampling of low resolution feature responses. SegNet [\ref{ref:17}] reuses the pooling indices from the encoder and learn extra convolutional layers to densify the feature responses. FCN and deeplabV3+ use the bilinear upsampling in the decoder module and connect the upsampling outputs with the low-level feature from the encoder module. DeeplabV2 [\ref{ref:11}] use Conditional Random Field (CRF) to improve the segmentation effect.

\subsection{Semantic segmentation for lane marking}
Azimi S. M.[\ref{ref:2}] use the aerial photography for lane marking semantic segmentation. This method can efficiently and quickly complete large-area lane marking semantic segmentation tasks. But there are two shortcomings, this method is easily interfered by the surrounding scenes in the picture and less robustness in areas where the illumination changes significantly. [\ref{ref:18}] combine the ConvLSTM with an encoder-decoder structure DCNN and using the time context information in the lane marking semantic segmentation task. This method has a certain improvement in the segmentation effect, but there are only two classification results, road and lane marking.

The semantic segmentation inputs of the above methods are all images come from cameras, and the segmentation results are difficult to meet the requirements of high-precision map construction as described above.
\section{The proposed method}\label{section:3}
In this section, we will first introduce the preprocessing of data, then we will introduce the proposed multi-sensor fusion deep neural network.

\subsection{data preprocessing}
Since the network contains two input branches, we will refer the input of the C-Region as the Branch C, and the input of the LBEV data as the Branch L. So we need to preprocess the data from the camera and LIDAR to meet the input requirements of the proposed network.
\begin{figure}[tp]
\centering
\begin{minipage}[t]{8.8cm}
\flushleft
\includegraphics[scale=0.2]{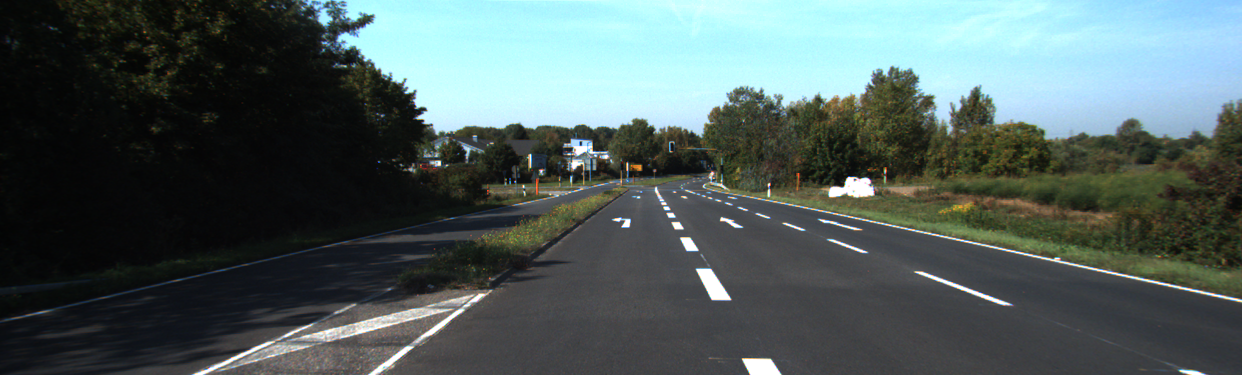}
\end{minipage}
\begin{spacing}{1.1}

\end{spacing}
\begin{minipage}[t]{10cm}
\flushleft
\includegraphics[scale=0.2044]{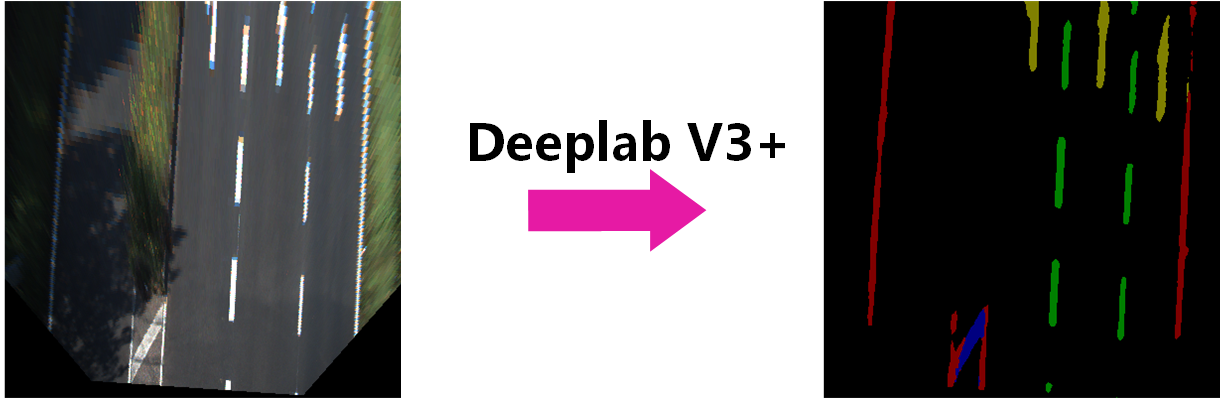}
\end{minipage}
\caption{The semantic segmentation result of CBEV (C-Region) acquisition process.}\label{fig:3}
\end{figure}
\subsubsection{Acquisition of C-Region}
First, we transform the front view of the camera into an bird's eye view through inverse perspective mapping. The CBEV is a 400 by 400 pixel image used to indicate the area of 26 meters to 6 meters in front and 10 meters from each side. That is to say, each pixel represents an area of 5 cm by 5 cm in real space. Then, we use the deeplabv3+ network trained on our labeled dataset to semantically segment it. So that we can get the input data for branch C, as show in Fig. 3. It should be noted that, C-Region is a single-channel grayscale image which is colored into an RGB image for convenience of observation in this paper.
\begin{figure*}[htp]
\centering
\includegraphics[scale=0.22]{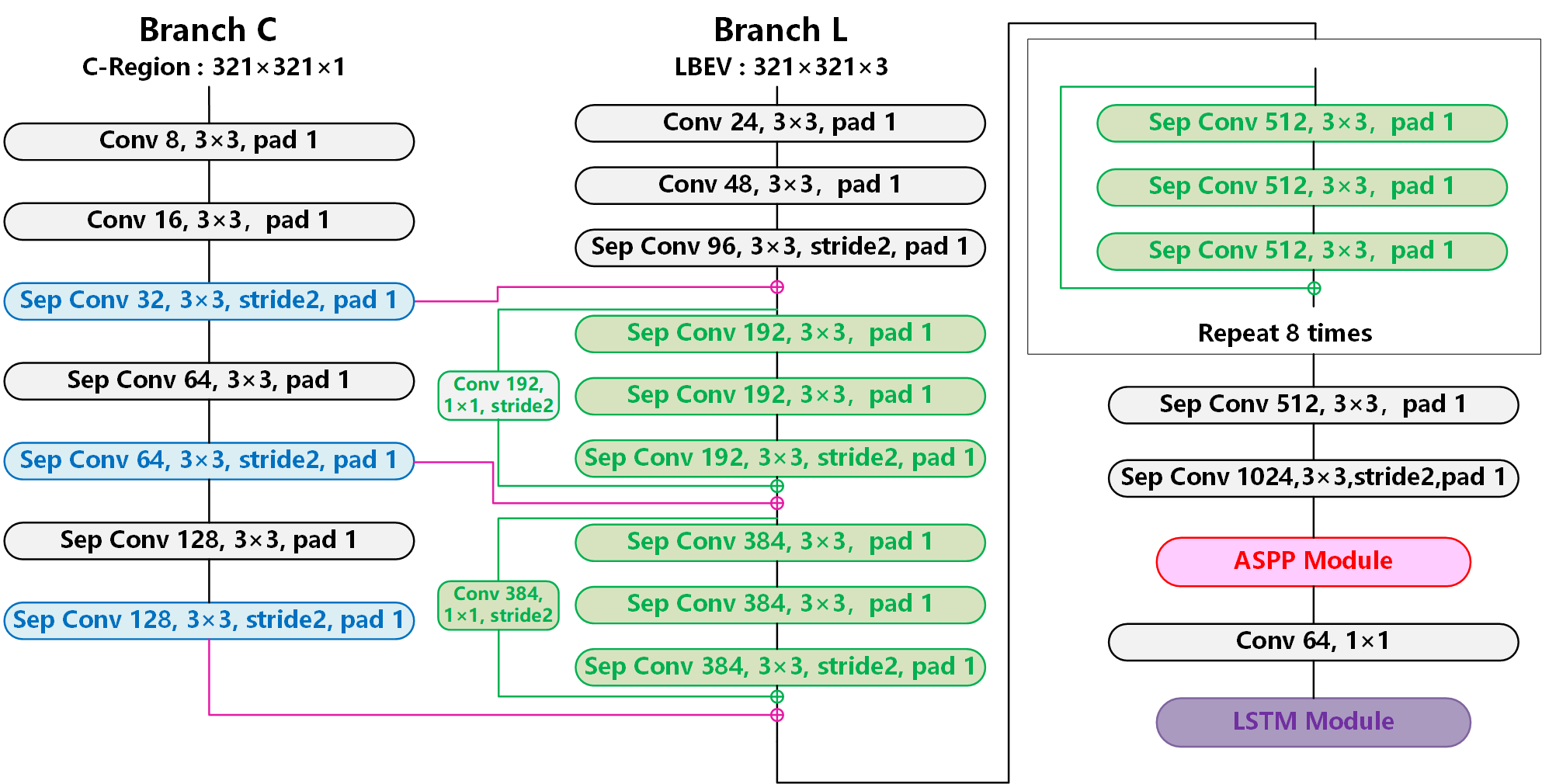}
\caption{The specific structure of the encoder module, the convolutional layers of different colors in the figure correspond to the different structures in Fig.  2. Limited by the size of the GPU memory, the images will be randomly cut to a size of 321¡Á 321 during the training.}\label{fig:4}
\end{figure*}
\subsubsection{LBEV design and generation}
We intercepted the same region of interest according to the CBEV for the original points cloud acquired by three-dimensional LIADR. According to the height information of the points cloud, the height threshold of the region of interest is between -2 meters and -1 meters (the installation height of the LIDAR is about 1.73 meters from the ground). In this paper, we are not simply projecting the points cloud into a two-dimensional grayscale image but transforming it into a three-channel bird's eye view, as shown in the second row of Fig. \ref{fig:1}.
The same as CBEV, each pixel of LBEV corresponds to a 5cm by 5cm real space. The value of the LBEV's first channel is corresponding to the intensity of the points spot falling within the grid. The specific calculation formula is as follows:
$$F(x,y)= \frac{\sum_{1}^{n}i}{n} \times  255\eqno{(1)}$$

where $F(x,y)$ is the value of the first channel, $ i_{1},i_{2},...,i_{n},i\in [0,1]$ is the reflection intensity value of each point falling within the grid which corresponding to the pixel, n is the number. The value of the second channel corresponds to the average height information of the points. The specific calculation formula is as follows:
$$S(x,y) = \frac{\sum_{1}^{n}(h+2)}{n} \times  255\eqno(2)$$

where $S(x,y)$ is the value of the second channel, $ h_{1},h_{2},...,h_{n},h\in [-2,-1]$  is the height value of each laser spot falling within the grid, $n$ is the number. The value of the third channel corresponds to the standard deviation of the height value of the points which falling within the grid and its eight neighborhoods. The specific calculation formula is as follows:
$$T(x,y) = 255 \times  \frac{2}{\pi} \times  \arctan \sqrt\frac{{\sum_{1}^{n}(h-\sum_{1}^{n}\frac{h}{n})^{2}}}{n}\eqno(3)$$

At the beginning, like the previous two channels, we set the statistical range of the standard deviation to a single pixel. But during the experiment, we found that a single pixel in LBEV often only corresponds to one or two LIDAR points, which makes the standard deviation meaningless. So we extend the statistical range to each pixel and its own eight neighborhoods. In formula (3), we use $arctan$ as the normalization function.

Through the above steps, we can get the input data for the Branch L.

\subsection{the proposed network}
Given the excellent performance\footnote[1]{http://host.robots.ox.ac.uk:8080/leaderboard/displaylb.php$?$challengeid=11
\&compid=6} of deeplabv3+ network on the PASCAL VOC 2012 semantic segmentation benchmark[\ref{ref:21}], the Xeception network shows greater performance than the ResNet. So in the proposed network, we chose the modified Xeception network as the backbone network. Here, we denote $output stride$ as the ratio of input image spatial resolution to the final output resolution.
\subsubsection{encoder module}
As shown in Fig. \ref{fig:4}, in the encoder module, we use two branches to perform convolution operations on LBEV and C-Region respectively. In the convolving process of C-Region, the output feature map will be  transmitted to the corresponding position of the branch L when its size is compressed to half of the input. At the very beginning, it wasn't clear to us how large the feature map should be when we taking the fusion operation. But in the end we decided to give the choice to the network itself, so there is a fusion operation each time when the feature map is compressed, and then the network can learn the best fusion strategy through the data. In this way, the network can learn the classification information from the Branch C.

It can be seen that the ratio of the number of the feature maps channels from two branches is 1: 3 in each fusion. This is to match the ratio between the original input C-Region and LBEV. In the Xception network, the ordinary convolutional layers are replaced by the depthwise separable convolutional layers which can greatly reduces the computational complexity of the network and improve performance to some extent[\ref{ref:19}]. Through this module, the size of the output feature map is 21¡Á 21¡Á 1024.

In the proposed method, we made the following two improvements: first, replace the maxpooling by a convolutional layer with a step size equal to two, second, perform batch normalization operations after each convolutional layer.
\subsubsection{ASPP module}
The full name of ASPP is Atrous Spatial Pyramid Pooling[\ref{ref:14}]. By paralleling multiple atrous convolutional layers with different atrous rates, ASPP module can help the network effectively captures multi-scale information. Just the same with DeeplabV3+, ASPP module in our network consists of one 1 ¡Á 1 convolution and three 3 ¡Á 3 convolutions with $atrous$  $rates$ = (6, 12, 18), and the image-level features. Each of them contains 256 channels, after connecting them in series, we will compress the thickness to 64 using a 1¡Á 1 convolutional layer, then the feather map will be entered into the LSTM module.
\begin{figure}[htp]
\centering
\includegraphics[scale=0.20]{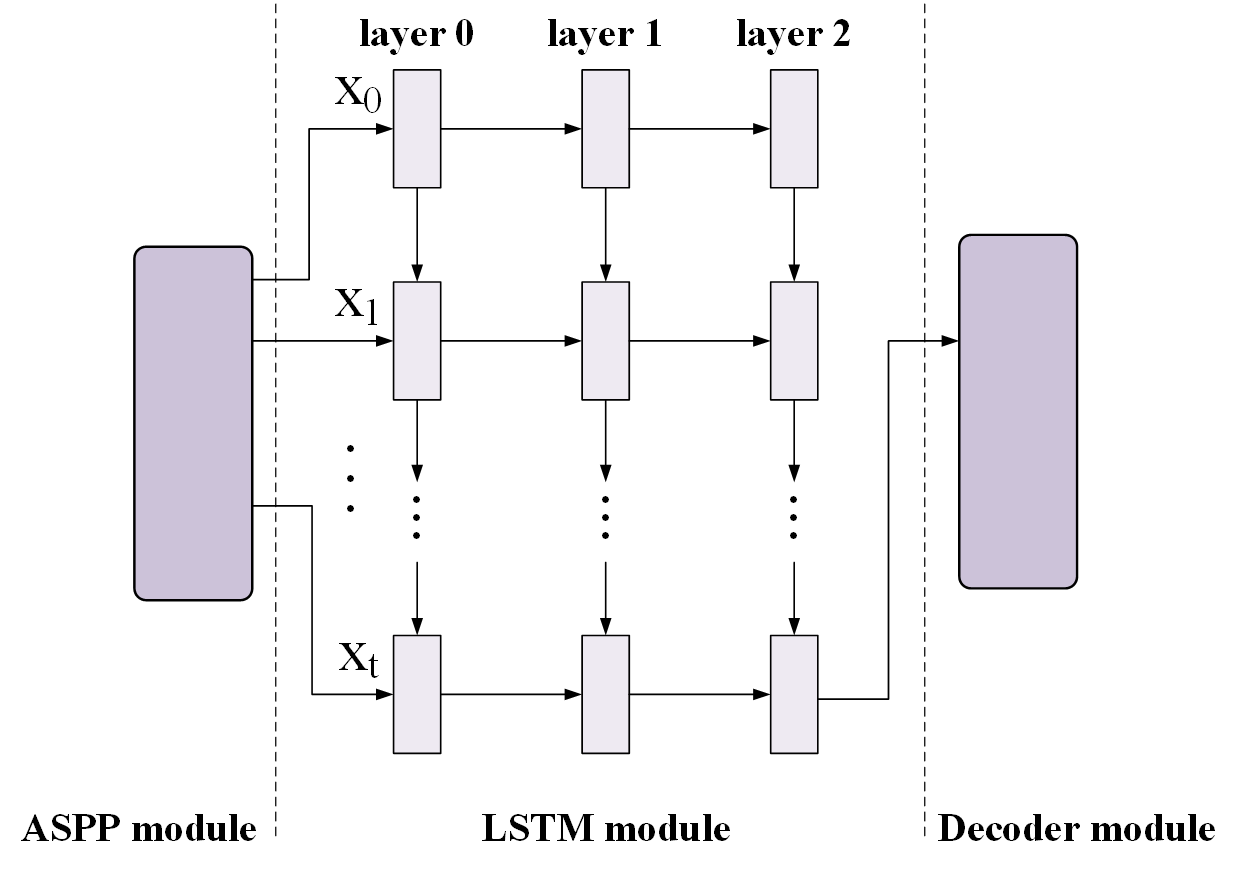}
\caption{The LSTM module, we also tuning the number of layers that should be included in the LSTM module, which was finally determined to be a one-layer structure.}\label{fig:5}
\end{figure}
\subsubsection{LSTM module}
In a real driving scenario, the acquired data of the sensor is continuous in time. It is easy to inspire us to input these continuous data into the RNN to help the network better perform the classification task. Specifically, an LSTM module is employed, which generally outperforms the traditional RNN model with its ability in forgetting unimportant information and remembering essential features. This module can also reduce the negative impact of errors in the C-Region to the network. The traditional full-connection LSTM is not only time- and computation- consuming but also cannot describe local features in the image, so a three-layer convolutional LSTM (ConvLSTM)[\ref{ref:20}] is applied in the proposed network instead of a full-connection LSTM structure, as shown in Fig. \ref{fig:5}.

The calculation process in a ConvLSTM cell can be formulated as:
\begin{spacing}{0.7}
\begin{align*}
i_{t}&=\sigma (W_{xi}*\mathcal{X}_{t}+W_{hi}*\mathcal{H}_{t-1}+W_{ci}\circ \mathcal{C}_{t-1}+b_{i})\\
f_{t}&=\sigma (W_{xf}*\mathcal{X}_{t}+W_{hf}*\mathcal{H}_{t-1}+W_{cf}\circ \mathcal{C}_{t-1}+b_{f})\\
o_{t}&=\sigma (W_{xo}*\mathcal{X}_{t}+W_{ho}*\mathcal{H}_{t-1}+W_{co}\circ \mathcal{C}_{t-1}+b_{o})\  \  \  \  \  \  (4)\\
\mathcal{C}_{t}&=f_{t}\circ \mathcal{C}_{t-1}+i_{t}\circ tanh(W_{xc}*\mathcal{X}_{t}+W_{hc}*\mathcal{H}_{t-1}+b_{c})\\
\mathcal{H}_{t}&=o_{t}\circ \mathcal{C}_{t}
\end{align*}
\end{spacing}
In ConvLSTM, the full-connection between each gate is replaced by a convolution operation. In the above formulas,  `*' and `$\circ$' denote the convolution
operation and the Hadamard product, respectively. $\mathcal{C}_{t}$, $i_{t}$, $f_{t}$ and $o_{t}$ represent the cell, input, forget and output gates. $\mathcal{C}_{t}$, $\mathcal{H}_{t}$, $\mathcal{C}_{t-1}$ and $\mathcal{H}_{t-1}$ represent the memory and output activations at time $t$ and $t-1$, respectively. $W_{xi}$ is the weight matrix of the input $\mathcal{X}_{t}$ to the input gate, $b_{i}$ is the bias of the input gate. The meaning of other
$W$ and $b$ can be inferred from the above rule. $\sigma$ represents the sigmoid operation and $tanh$ represents the hyperbolic tangent non-linearities.

\subsubsection{decoder module}
\begin{figure}[htp]
\centering
\includegraphics[scale=0.255]{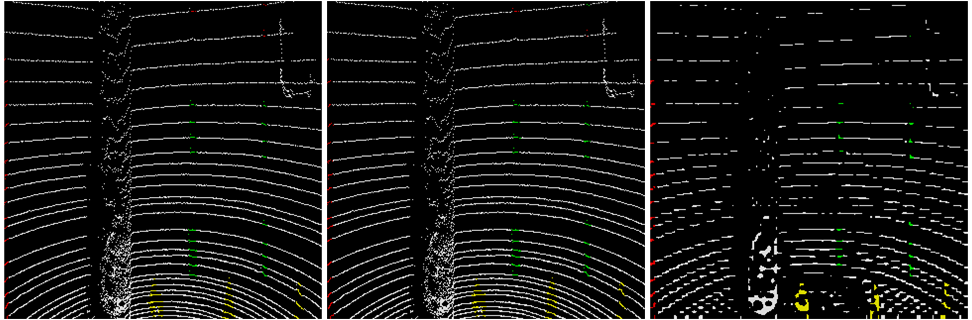}
\caption{The Comparison of different decoder  structures. Left: the ground-truth. Middle: the decoder with two times of low-level feature fusion. Right: the decoder with only one low-level feature fusion.}\label{fig:6}
\end{figure}

In the decoder module, the feature image output by the LSTM module will will be restored to the same size as the original image after two times of bilinear upsampling. In the beginning, we only fuse the low-level feature from the encoder module during the first upsampling process. But the results show that the network cannot accurately recover the details of the original image. So we directly merge the original input image into the second upsampling process after a 1x1 convolution operation, and the result shows that it greatly improves the detail recovery effect of the image, as shown in Fig. \ref{fig:6}. It should be noted that all low-level feature comes from the Branch L instead of the Branch C. In our decoder module,
\subsubsection{Training strategy}
In the proposed method, we use Momentum[\ref{ref:22}] optimizer first, however it is not suitable for the lack of reliable initialization parameters. And the Momentum optimizer is insufficiently convergent on our dataset from the performance during the training. So we use Adam[\ref{ref:23}] optimizer instead. In the tuning process, we mainly consider the following aspects.
\begin{itemize}
 \item[¡¤]In the aspect of network structure, we tried the encoder module with different number of ResBlocks and the LSTM module with different number of ConvLSTM layers.
 \item[¡¤]We tuning the time step in the LSTM module and the batch size of the input data and the learning rate and its decay during the training.
 \item[¡¤]To overcome the imbalance among the classes, we chose a weighted cross entropy as the loss function and tuning the weights of the different samples.
\end{itemize}
\section{Experiment and results}\label{section:4}
In this section, the datasets used in this paper will be firstly introduced. Then the experiments are conducted to verify the validity and accuracy of the proposed method. The performances of the proposed method which evaluated in the datasets are compared with the state of the art methods in the field of semantic segmentation.
\begin{figure}[htp]
\centering
\includegraphics[scale=0.25]{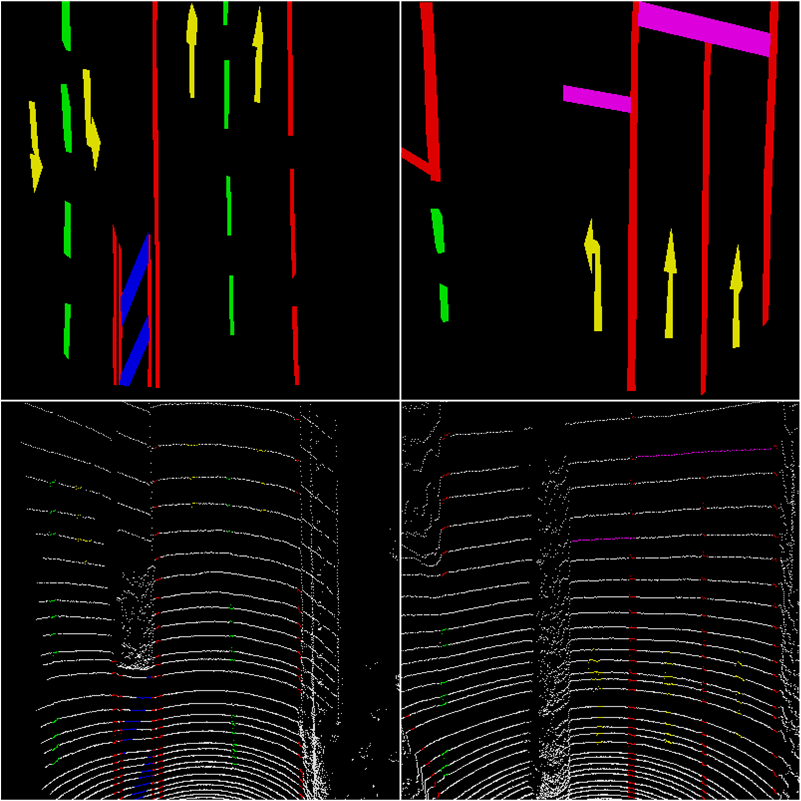}
\caption{The labeled images. The first row is labeled CBEV, black area is the Background, red area is the Solid Line, green area is the Dotted Line, purple area is the Stop Line, yellow area is the Arrow and the blue area is the Prohibited Area. The second row is labeled LBEV, added a white area representing the Other Point.}\label{fig:7}
\end{figure}
\subsection{Datasets}
We construct our own datasets based on KITTI darasets because it contains synchronous and continuous images and point cloud data of the road. We manual labeled 436 images of the LBEV and CBEV each.

It should be noted that we take the 81th to 148th images as the testing set. Then we rotate the picture 20 times both clockwise and counterclockwise, one degree each time. So we get a datasets with 14720 labeled LBEV and CBEV each. It ensures that the objectivity of the testing set. We use the rotated images as the training set which has a number of $362 \times  40 = 14480$, then we use the original 362 images as the validation set. During training, we will test the model on the validation set when each epoch is finished. With Tensorboard¡ªa tool provided by Tensorflow$[\ref{ref:24}]$¡ªwe can track the performance of the model on the training and validation sets in real time, so as to adjust the training strategy in time.

In our datasets, we divide the CBEV into six different labels which are Background, Solid Line, Dotted Line, Stop Line, Arrow and Prohibited Area, and for the LBEV, We added an additional category called Other Point, as shown in Fig. \ref{fig:7}. We use the labeled CBEV to train DeeplabV3+ based on the cityscapes pretrained model\footnote[2]{http://download.tensorflow.org/models/deeplabv3\_cityscapes\_train\_2018\_
02\_06.tar.gz}.

Then we use the labeled LBEV and the C-Region predicted by DeeplabV3+ to train our own network.

\subsection{Experimental platform}
The experiments are implemented on a computer with an Intel Core i7-8700@3.2GHz, 32GB RAM and one NVIDIA TITAN-X (Pscal) GPU.
\begin{table}[htp]
\centering
\caption{MIOUs of of the two optimizers on the testing set}
\label{table1}
\begin{tabular}{@{}ccc@{}}
\toprule
Optimizer & Momentum & ADAM \\ \midrule
MIOU(\%) & 43.58 & 67.43 \\ \bottomrule
\end{tabular}
\end{table}
\begin{figure}[htp]
\centering
\includegraphics[scale=0.18]{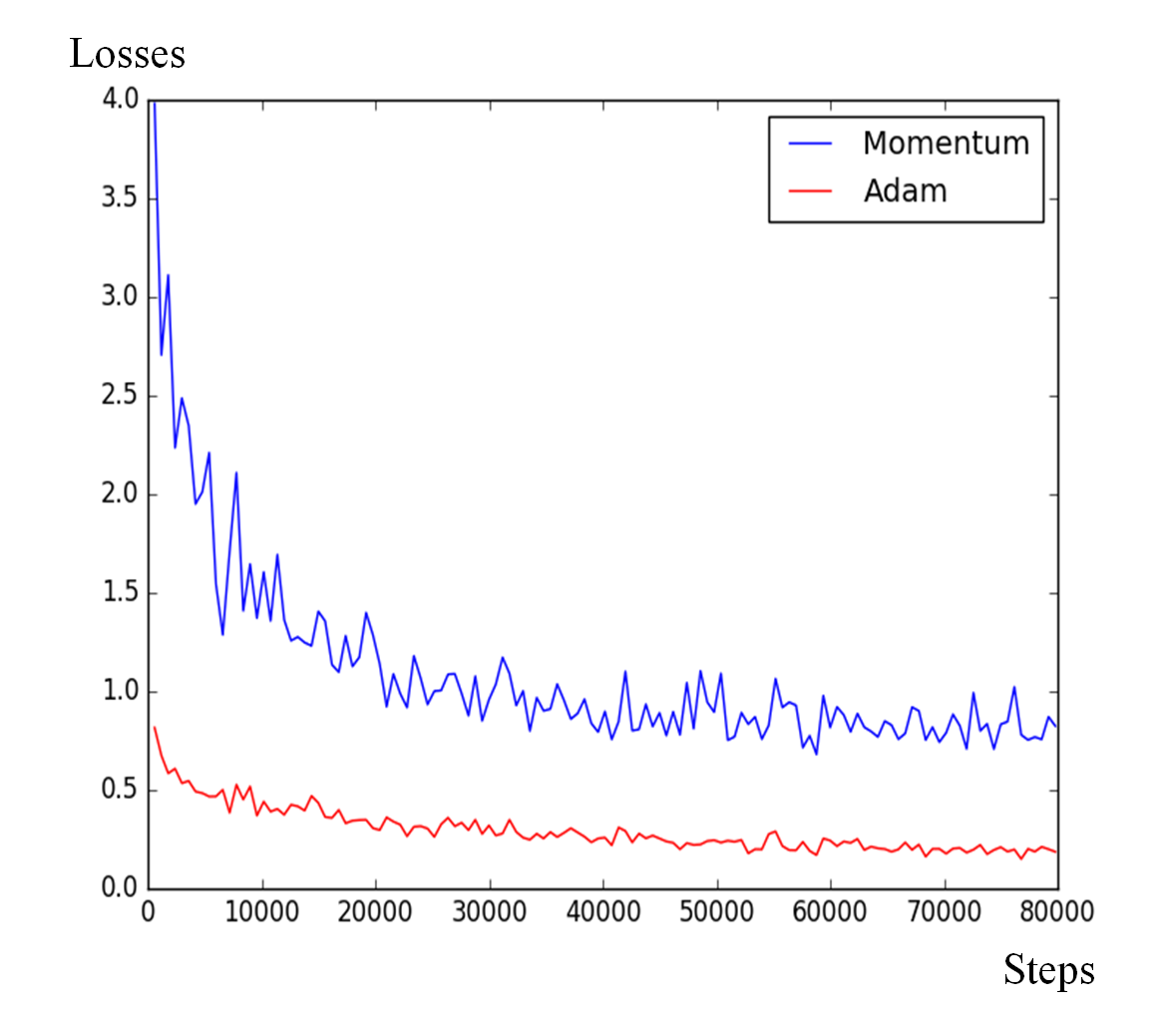}
\caption{The losses of two optimizers during training.}\label{fig:8}
\end{figure}
\subsection{Transfer learning for DeeplabV3+}
As mentioned above, we train DeeplabV3+ on our CBEV training set based on the pretrained model. During the training process, we found that the Momentum optimizer in DeeplabV3+ could not converge well on the training set, so we used the ADAM optimizer instead. Fig. \ref{fig:8} and Table \ref{table1} shows the final losses on our CBEV training set and the MIOU on the testing set of these two different models, respectively. It can be seen that the DeeplabV3+ model with Adam optimizer can achieve better semantic segmentation results of CBEV. So the C-region we need comes from the prediction results of the improved DeeplabV3+ model.

\subsection{Performance and comparison on LBEV semantic segmentation}
This section mainly contains two types of comparative on the experiment results. First, we compare the segmentation results of each model intuitively and visually. Second, we will quantitatively analyze and compare each model. In these experiments, we mainly compare our proposed model with the DeeplabV3+ model which achieved excellent results on the PASCAL VOC 2012 semantic segmentation benchmark.
 \begin{figure*}[htp]
\centering
\includegraphics[scale=0.225]{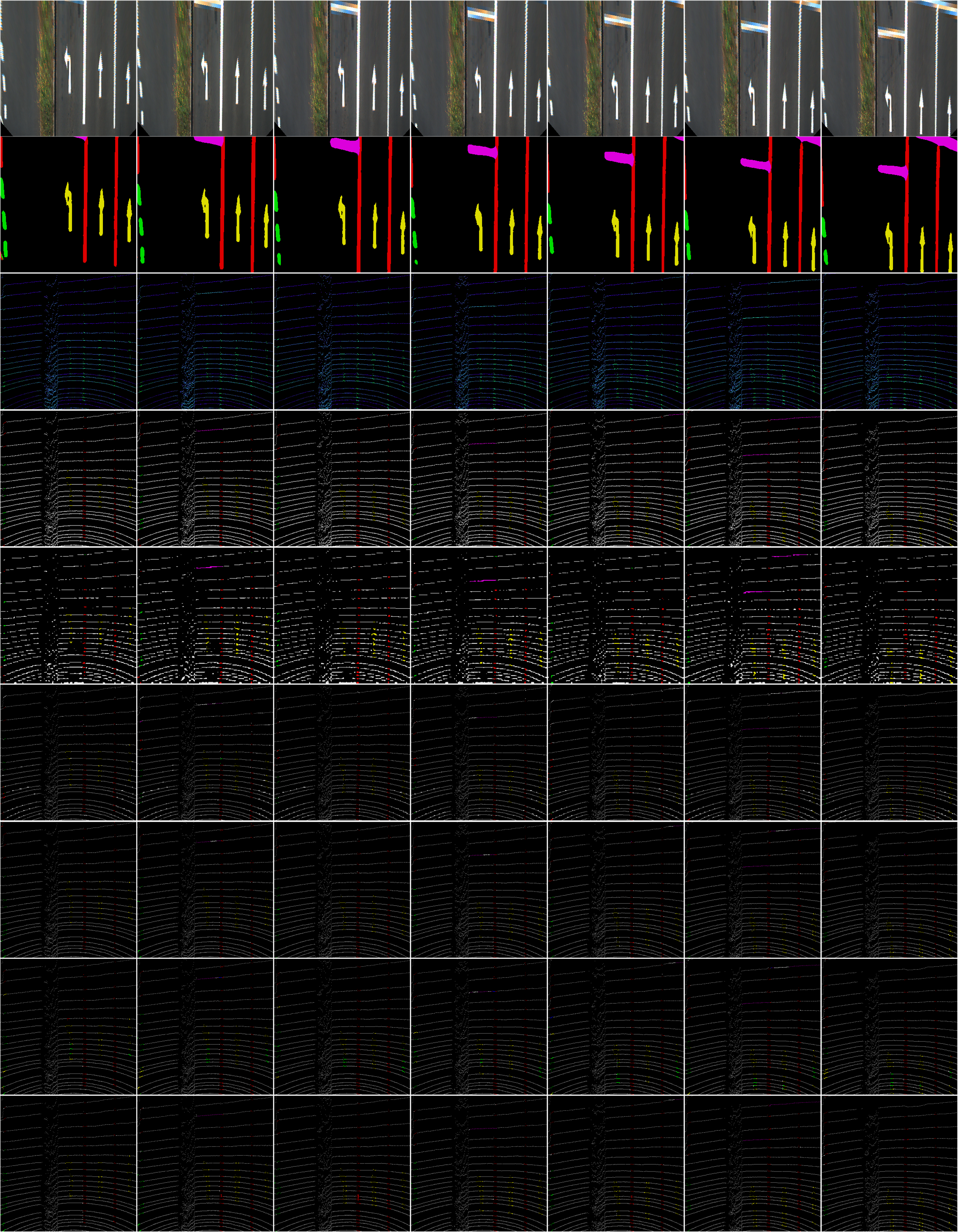}
\caption{Raw data and segmentation results of different models in seven consecutive scenarios. First row, the CBEV. Second row, the C-Region obtained from DeeplabV3+. Third row, the LBEV. Fourth row, the ground-truth. Fifth row, DeeplabV3+. Sixth row, the Modified DeeplabV3+. Seventh row, the FusionLane\_Without\_LSTM. Eighth row, the FusionLane\_FcLSTM. Ninth row, the FusionLane. (For the last four columns of images, we reduce the brightness of the correctly classified pixels and highlight the incorrectly classified pixels. Zoom in to see it more clearly. The original images are available at \href{https://drive.google.com/open?id=1iKv0c2A6UXud\_HzWUPXIPUsqTQhvdcey}{$here$}.)}\label{fig:9}
\end{figure*}

Specifically, in the experiment we compare the following models.
 \begin{itemize}
  \item[¡¤]DeeplabV3+: Developed by Google, it is one of the most advanced model of image semantic segmentation. We perform LBEV and CBEV semantic segmentation experiments on it respectively.
  \item[¡¤]Modified DeeplabV3+: Considering the particularity of this task, we made two improvements based on DeeplabV3+. First, we apply a 1¡Á 1 convolution on the original input LBEV, then concatenate it with the output of the second upsampling. Second, replace the Momentum optimizer with the Adam optimizer.
  \item[¡¤]FusionLane\_Without\_LSTM: This model does not contain LSTM module but all other structures are the same as the description in Section III.
  \item[¡¤]FusionLane\_FcLSTM: We deploy a traditional full-connection LSTM after the ASPP module.
  \item[¡¤]FusionLane: It is exactly the model proposed in this paper which contains a ConvLSTM structure.
 \end{itemize}

\subsubsection{Visually intuitive evaluation}
 The segmentation results on the testing set obtained by the above models after training are shown in Fig. \ref{fig:9}.

In Fig. \ref{fig:9}, for the visually intuitive comparison we have selected a set of segmentation results that contain seven consecutive scenes. The first four rows are CBEV, C-Region, LBEV and ground-truth of LBEV, respectively. The fifth row is the prediction results of the DeeplabV3+, we can see that the decoder is difficult to restore the LBEV perfectly because the original input image is not merged during the upsampling process and lost a lot of details, resulting in a very low MIOU.

From the sixth line, we can see that through our improvements, the Modified DeeplabV3+ can recover the details of the original image quite well but not perfect, meanwhile,  due to the lack of classification information from the C-Region, the network can only rely on the features in the LBEV and is likely to produce incorrect classification results. For example, in this row, the model struggles with predictions of solid and dotted lines in the left half of the images because there is a quite large distance between the two laser lines and it is in an intersection scene where the solid line and the dotted line are very close. At the same time, the model's prediction of the stop line is also bad.

The seventh line is the segmentation result of FusionLane\_Without\_LSTM, this model can be seen as combining the classification information of C-Region on the basis of the above model. It performs well in most cases, but mistake happens when there are serious classification error or blind spot in the C-Region. There happens to be no serious errors in C-Region from the enumerated scene but it happens quite often. We can still find some errors in the predictions of the dotted line and stop line.

We hope to overcome the above problems by using the timing information between the frames before and after. So we deploy a traditional full-connection LSTM in the model which been called as FusionLane\_FcLSTM. However, from the segmentation results in the eighth row, the negative impact of the FcLSTM structure is greater than the positive impact. In this row, we found that the model produces many outrageous errors, such as confusing arrows with dotted lines and so on. This is mainly because the feature map is transformed into a one-dimensional tensor when input to the FcLSTM module, which destroys the local features that existed.

From the above analysis, in order to complete the LBEV semantic segmentation task well, Our model improvement ideas are divided into the following three steps:
\begin{itemize}
 \item[¡¤]First, it is difficult for the network to obtain excellent semantic segmentation results based on the information provided by LBEV only. We need to introduce the C-Region.
 \item[¡¤]Second, error messages and blind spots in the C-Region can have a negative impact on the semantic segmentation results. We can overcome this problem through timing information.
 \item[¡¤]Third, the FcLSTM module destroys local features in the feature maps, resulting in new errors. So we replace the FCLSTM module with a ConvLSTM module.
\end{itemize}

From the last row we can see that the FusionLane model completed the task very well and got a nearly perfect semantic segmentation result.

\subsubsection{Quantitative evaluation}
In the quantitative evaluation, we mainly compare the accuracy, Intersection over Union on each category (IOU), the Mean-IOU (MIOU) and Pixel Accuracy of different models on the testing set. As shown in Table \ref{table2}.

\begin{table*}[htp]
\caption{IOU on each category, MIOU and the Pixel Accuracy of different models}
\label{table2}
\resizebox{\textwidth}{!}{
\begin{tabular}{@{}lccccccccc@{}}
\toprule
\multicolumn{1}{c}{Methods} & Background & Solid Line & Dotted Line & Arrow & Prohibited Area & Stop Line & Other Point & MIOU & Pixel Accuracy(\%) \\ \midrule
DeeplabV3+ (LBEV) & 0.9419 & 0.2587 & 0.2648 & 0.2793 & 0.1915 & 0.3586 & 0.2770 & 0.3674 & 91.31 \\
DeeplabV3+ (CBEV) & 0.9106 & 0.6287 & 0.7012 & 0.5821 & 0.6935 & 0.5294 & - & 0.6743 & 85.76 \\
Modified DeeplabV3+ & 0.9989 & 0.6480 & 0.6230 & 0.7106 & 0.5788 & 0.3024 & 0.9654 & 0.6896 & 99.59 \\
FusionLane\_Without\_LSTM & \textbf{1.0000} &  0.7285 & 0.7752 & \textbf{0.7653} & 0.7426 & 0.6574 & 0.9864 & 0.8079 & 99.87 \\
FusionLane\_FcLSTM & 0.9999 & 0.7004 & 0.6192 & 0.5491 & 0.6830 & 0.5629 & 0.9838 & 0.7283 & 99.81 \\
FusionLane & \textbf{1.0000} & \textbf{0.7477} & \textbf{0.7838} & 0.7526 & \textbf{0.7979} & \textbf{0.9053} & \textbf{0.9867} & \textbf{0.8535} & \textbf{99.92} \\ \bottomrule
\end{tabular}}
\end{table*}

Pixel Accuracy and MIOU are very common evaluation indicators in the field of semantic segmentation, they can be calculated as:
$$PixelAccuracy=\frac{\sum_{i}n_{ii}}{\sum_{i}t_{i}}\eqno(5)$$
$$MIOU=\frac{1}{n_{c}}\sum_{i}\frac{n_{ii}}{(t_{i}+\sum_{j}n_{ji}-n_{ii})}\eqno(6)$$

where $n_{c}$ is the number of classes included in ground truth segmentation, $n_{ij}$  denotes the number of pixels of class $i$ predicted to belong to class $j$ and $t_{i}$ is the total number of pixels of class $i$ in ground truth segmentation.

From Table  \ref{table2} we can find out DeeplabV3+ is totally unsuitable for LBEV semantic segmentation task. The IOU of all classes except Background are very low. But this is not because of inadequate training. Because when we made some modifications to the structure, the Modified DeeplabV3+ make a breakthrough under the same training strategy. For CBEV semantic segmentation task, DeeplabV3+ achieved much better result than the LBEV task, the MIOU is 67.43\%, but still not good enough. These three lines of data show that the single-sensor-based approaches do not perform well in this task.

Because in this paper, we are focus on the LBEV semantic segmentation task, we use the the Modified DeeplabV3+ to replace the original DeeplabV3+ for comparison.

After merging the classification information of the C-Region, the performance of FusionLane\_Without\_LSTM is much better than the Modified DeeplabV3+  even if the network is smaller, the MIOU increased almost 12\% compared to Modified DeeplabV3+ and even achieved the best score on the IOU of Arrow class.

Compared to the previous model, the performance of FusionLane\_FcLSTM is the overall decline, this is the consequence of destroying local features.

From the last row in the table, we can see that the FusionLane model has achieved the best results in all indicators except for the IOU of Arrow, the MIOU is 16.39\% higher than the Modified DeeplabV3+ and also increased by 4.56\% compared with FusionLane\_Without\_LSTM.

From the above data, it is not difficult to see that if we rely on a single kind of sensor, whether camera or LIDAR, we cannot obtain sufficiently accurate semantic segmentation results. Effective fusion of data from different sensors is the key to solving the problem.

We found an interesting and confusing phenomena from the data in the table. In the first row, the IOU are very low in almost all categories except for the Background class but the Pixel Accuracy is relatively high which is 91.31\%. This is because, in addition to the Background class the $n_{ii}$ of all the other classes are very small which leads to two consequences. First, for these classes with a small percentage, a few misclassifications will cause the IOU to drop dramatically. Also means that it is difficult to improve the IOU when reaching a certain height. Second, for the Background class which occupies most of the LBEV, the performance of the models in this class largely determines the value of Pixel Accuracy. This not only eliminates our previous confusion but also can explains why DeeplabV3+ (LBEV) is lower in MIOU and higher in Pixel Accuracy compare with DeeplabV3+ (CBEV).

\subsubsection{Important parameter analysis}
Among all the hyperparameters, $time$ $step$ is one of the parameters that have the greatest influence on the network performance. It represents how many frames of historical data the network can use to help predict the current frame.

When the $time$ $step$ is large, the network can review more historical frames which means there are more historical information. But this does not mean that the larger the $time$ $step$ , the better the prediction of the network. Because when the $time$ $step$ is too large, some data in the history frame is likely to be significantly different from the current data, which will have a negative effect on the network prediction results. Therefore, we conducted a comparative experiment on the network performance with different $time$ $step$ values, as shown in Table \ref{table3}.

At the beginning, MIOU increased with the increase of $time$ $step$, which proves that historical information has a positive impact on the final result. The MIOU reached its peak when $time$ $step$ is 4, but after that, further increase of the $time$ $step$ will decrease MIOU, this phenomenon is completely consistent with the previous analysis.

\begin{table}[]
\centering
\caption{MIOU with different Time Step Values}
\label{table3}
\begin{tabular}{@{}cccccc@{}}
\toprule
Time Step & 2 & 3 & 4 & 5 & 6 \\ \midrule
MIOU(\%) & 81.73 & 83.52 & \textbf{85.35} & 81.86 & 82.14 \\ \bottomrule
\end{tabular}
\end{table}

\section{Conclusion}\label{section:5}
In this paper, we propose a semantic segmentation network for the LIDAR points cloud bird's eye views (LBEV) for the first time. Thanks to the accuracy of LBEV, the semantic segmentation results can be directly used to construct high-precision map.

For the task of lane marking semantic segmentation, the proposed method is not simply to turn it into a binary classification task but to further subdivide it into a multi-classification task.

From the network structure, we propose a network with a dual-input branch structure to fuse LEBV and C-Region. For the possible misclassification information in C-Region, we reduce the negative impact on network prediction results by adding LSTM structure to the network. Experiments show that the proposed method can effectively fuse information from LIDAR and camera, and achieve excellent results on LBEV semantic segmentation task.

In the future, we will work to transform the Branch C input from C-Region to CBEV to build an end-to-end semantic segmentation network and focus on getting more training data. And based on this, we will start the construction of high-precision maps.

\section{Acknowledgement}\label{section:6}
This work was supported by National Key Research and Development Program of China (Nos. 2016YFD0701401, 2017YFD0700303 and 2018YFD0700602), Youth Innovation Promotion Association of the Chinese Academy of Sciences (Grant No. 2017488), Key Supported Project in the Thirteenth Five-year Plan of Hefei Institutes of Physical Science, Chinese Academy of Sciences (Grant No.KP-2017-35, KP-2017-13,KP-2019-16), Independent Research Project of Research Institute of Robotics and Intelligent Manufacturing Innovation, Chinese Academy of Sciences (Grant No. C2018005), and Technological Innovation Project for New Energy and Intelligent Networked Automobile Industry of Anhui Province.

\end{document}